\documentclass{article}

\usepackage{arxiv}

\usepackage[utf8]{inputenc} % allow utf-8 input
\usepackage[T1]{fontenc}    % use 8-bit T1 fonts
\usepackage{hyperref}       % hyperlinks
\usepackage{url}            % simple URL typesetting
\usepackage{booktabs}       % professional-quality tables
\usepackage{amsfonts}       % blackboard math symbols
\usepackage{nicefrac}       % compact symbols for 1/2, etc.
\usepackage{microtype}      % microtypography
\usepackage{lipsum}		% Can be removed after putting your text content
\usepackage{graphicx}
\usepackage{cite}
\usepackage{doi}
\usepackage{xcolor}
\usepackage{array}
\newcolumntype{C}[1]{>{\centering\let\newline\\\arraybackslash\hspace{0pt}}m{#1}}

\title{Pellet-based 3D Printing of Soft Thermoplastic Elastomeric Membranes for Soft Robotic Applications}

\author{ Nick Willemstein \\
	Department of Biomechanical Engineering\\
        University of Twente\\
        Enschede, The Netherlands \\
	\texttt{n.willemstein@utwente.nl} \\
	\And
	Mohammad Ebrahim Imanian \\
	Department of Biomechanical Engineering\\
        University of Twente\\
        Enschede, The Netherlands \\
	\texttt{m.e.imanian@utwente.nl} \\
   	\And
	Herman van der Kooij \\
	Department of Biomechanical Engineering\\
        University of Twente\\
        Enschede, The Netherlands \\
	\texttt{h.vanderkooij@utwente.nl} \\
        \And
	Ali Sadeghi \\
	Department of Biomechanical Engineering\\
        University of Twente\\
        Enschede, The Netherlands \\
	\texttt{a.sadeghi@utwente.nl} \\
}

\renewcommand{\headeright}{}
\renewcommand{\undertitle}{}
\renewcommand{\shorttitle}{}

\begin{document}
\maketitle

\begin{abstract}
Additive manufacturing (AM) is a promising solution for handling the complexity of fabricating soft robots. However, the AM of hyperelastic materials is still challenging with a limited material range. Within this work, pellet-based 3D printing of very soft thermoplastic elastomers (TPEs) was explored (down to Shore Hardness 00-30). Our results show that TPEs can have similar engineering stress and maximum elongation as Ecoflex 00-10. In addition, we 3D-printed airtight thin TPE membranes (0.2-1.2 mm), which could inflate up to a stretch of 1320\%. Combining the membrane’s large expansion and softness with the 3D printing of hollow structures simplified the design of a bending actuator that can bend 180 degrees and reach a blocked force of 238 times its weight. In addition, by 3D printing TPE pellets and rigid filaments, the soft membrane could grasp objects by enveloping an object or as a sensorized sucker, which relied on the TPE’s softness to conform to the object or act as a seal. In addition, the sucker’s membrane acted as a tactile sensor to detect an object before adhesion. These results suggest the feasibility of AM of soft robots using soft TPEs and membranes as a promising material type and sensorized actuators, respectively.
\end{abstract}

\section{Introduction}
Soft robots are known for their ability to interact safely with both their users and their environment. Made from compliant materials and structures, they demonstrate high adaptability. Over the past decade, researchers have utilized a broad range of compliant materials, including rubber, textile, and paper, to fabricate soft actuators, sensors, and robots. Their results led to soft systems that incorporated mechanical compliance, sensing, and actuation in a single structure similar to the integration found in biological systems.

Like their biological counterparts, researchers often rely on composite structures to realize the desired functionality of soft robotic systems. These include composites of materials with different stiffnesses, such as combining rubber with fabric/fiber for soft fluidic actuators \cite{mosadegh2014pneumatic,connolly2015mechanical}. Whereas by adding stimuli-responsive materials to rubber, such as conductive textiles, researchers made soft sensors \cite{chiara, osborn2018prosthesis}.To fabricate these multi-material composites, fabrication methods that can handle and assemble multiple materials are essential.

This need makes additive manufacturing (AM) a promising approach, as it can fabricate complex shapes, access the inside of the structure, and incorporate multiple materials. For instance, the inkjet printing method \cite{buchner2023vision} realized soft robots with fluidic actuation and sensing capabilities by combining materials with a broad range of stiffnesses and benefiting from access to the inside of the structure. Another notable example is embedded 3D printing \cite{wehner2016integrated}, which combined multi-material 3D printing, molding, and soft lithography to fabricate a soft robot with integrated control, actuation, and power. Their approach relied on a mold to 3D print the soft material. Other researchers combined molding and 3D printing by printing an interlocking pattern, which was filled by casting silicone rubber \cite{goshtasbi2025bio,rossing2020bonding}. However, the need for a mold in both cases limited the design freedom, which would be avoided in direct 3D printing strategies. 

Unfortunately, direct 3D printing of hyperelastic soft materials is still challenging, which researchers are trying to overcome by developing new systems, materials, and strategies. A notable example of 3D printing hyperelastic materials is ink-based material extrusion (MEX) of silicone rubbers \cite{morrow2016directly,walker2019zero}, which can achieve performance similar to molded samples \cite{mold}. In addition, ink-based MEX has a wide range of materials \cite{liu2025advances}, including the capability to fabricate soft robots with soft and rigid materials \cite{voxelated}. 

Another notable class of AM is the filament and pellet-based Material Extrusion (MEX) processes, which primarily use thermoplastics. Thermoplastics have the advantage of fast solidification and can be thermally reprocessed. Previous works have 3D-printed PneuNet actuators \cite{yap2016high,tawk20223d} and a sensorized hybrid hand \cite{grignaffini2024new} that incorporated flexible filament with a Shore Hardness of $>$70A using filament-based MEX. 

In general, filament-based MEX processes extrude material by pushing a filament through a heated nozzle. This pushing can lead to buckling for very soft filaments, which restricts the Shore hardness within filament-based MEX printers to $>$60A \cite{liu2025advances}. Pellet-based MEX processes circumvent the stiffness requirement by dragging pellets with a screw instead. 

This dragging-based approach enables pellet extruders to directly 3D print thermoplastic elastomers (TPEs), which can be significantly softer than flexible filaments. Researchers have investigated soft TPEs (Shore Hardness $<$50A), such as styrene-ethylene-butylene-styrene (SEBS), for soft fluidic actuators. For instance, in \cite{khondoker2019direct}, a PneuNet actuator using a TPE with Shore Hardness 47A was successfully printed. Other researchers showed a two-stage printing approach using SEBS with and without carbon black to fabricate a sensorized PneuNet actuator \cite{Georgopolou}. Furthermore, by exploiting the behavior of the material during deposition, researchers could print porous structures made from conductive TPE for sensorized vacuum actuators \cite{WillemsteinPRA}. 

Pellet-based MEX has also been shown to print other useful materials for soft robotics, such as those with very high elongation (4048\%) \cite{bayati20243d} and nano-composites with magneto-thermal shape memory-capabilities \cite{mirasadi20243d}. In addition, researchers have shown the printability of bio-degradable polymers using pellet extruders \cite{karimi2024direct} and the ability to scale to larger prints \cite{goh2024large}. These capabilities indicate that pellet-based MEX is an active field of research, which can help meet the needs of fabricating soft robots. 

For soft robotics, the Shore hardness is an important metric. The printed Shore Hardness in pellet-based MEX processes for soft robotics is, however, still relatively stiff, and includes 18A \cite{Georgopolou}, 47A \cite{khondoker2019direct}, and 70A \cite{WillemsteinPRA}. Thus, although pellet-based MEX demonstrated promising results in sensing, actuation, and sensorized actuators, the pellets are much stiffer than soft silicone rubbers such as the Eco-flex family, which can reach Shore Hardness 00-10. Crossing into the 00-range would enable pellet-based MEX processes to reach silicone rubber-level stiffness with the processing advantages of thermoplastics (mentioned earlier). The first promising results of using such soft TPEs have been seen in \cite{curmi2025screw}. However, we focus on utilizing a very soft TPE (00-30) and a slightly stiffer TPE (47A) to realize soft fluidic (sensorized) actuators.

More specifically, we explore the potential of pellet-based MEX to fabricate soft robotic components by directly 3D printing TPE membranes and integrating them into functional (sensorized) fluidic actuators. Our main contribution lies not in the hardware of the pellet extruder itself, which is kept minimal and low-cost, but in demonstrating how TPE membranes, printed with pellet extrusion, can replace geometrically complex designs in soft robots. 

We validated this capability through the design and characterization of three functional membrane-based demonstrators: a membrane-based bending actuator, a suction cup, and a gripper. These demonstrators will show that a softness and performance comparable to established systems (e.g., PneuNets \cite{yap2016high,khondoker2019direct}) can be achieved without geometrical tricks, through the soft TPEs and pellet-based MEX process. 

In addition, we demonstrate that a 3D-printed membrane-based sucker can integrate sensing and self-sealing through the soft and airtight membranes combined with a rigid core, which is a novel way of utilizing the membrane of these suckers as well. By slightly modifying this design, another type of gripper can be realized that benefits the membrane and softness for grasping objects by conforming and providing high friction. Lastly, all these demonstrators benefit from both the softness and 3D printing's ability to print hollow structures. 

Besides the demonstrators, we characterized the 3D-printed TPE uniaxial stiffness and the inflation behavior of the TPE membranes. These experiments provide insight into the stretching and the effect of design (thickness, infill pattern) and control (pressure magnitude) aspects. 

In addition, a secondary contribution of this paper is to show our pellet extruder prototype, which incorporates a simple cooling system to prevent caking of the pellets.

This paper is organized as follows: First, our pellet extruder design and the TPE’s uniaxial properties are discussed. Afterward, we characterize the membrane's stretching behavior during inflation. These membranes are then used to 3D print three demonstrators, namely (in order): a bending actuator, a sensorized sucker, and a membrane-based gripper. Subsequently, a discussion and conclusion contextualize our results.

%%%%%%%%%%%%%%%%%%%%%%%%%%%%%%%%%%%%%%%%%%%%%%%%%%%%%%%%%%%%%%%%%%%%%%%%%%%%%%%%%%%%%%%%%%%%%%%%%%%%%%%%%%%%%%%%%%%%%%%%%%%%%%%%%%%%%%%%%%%%%%%%%%%%%%%%%%%%%%%

\section{Results}

\subsection{Pellet Extruders for Printing Very Soft Thermoplastic Elastomer Membranes}

\begin{figure}
\centering
\includegraphics[width=0.98\textwidth]{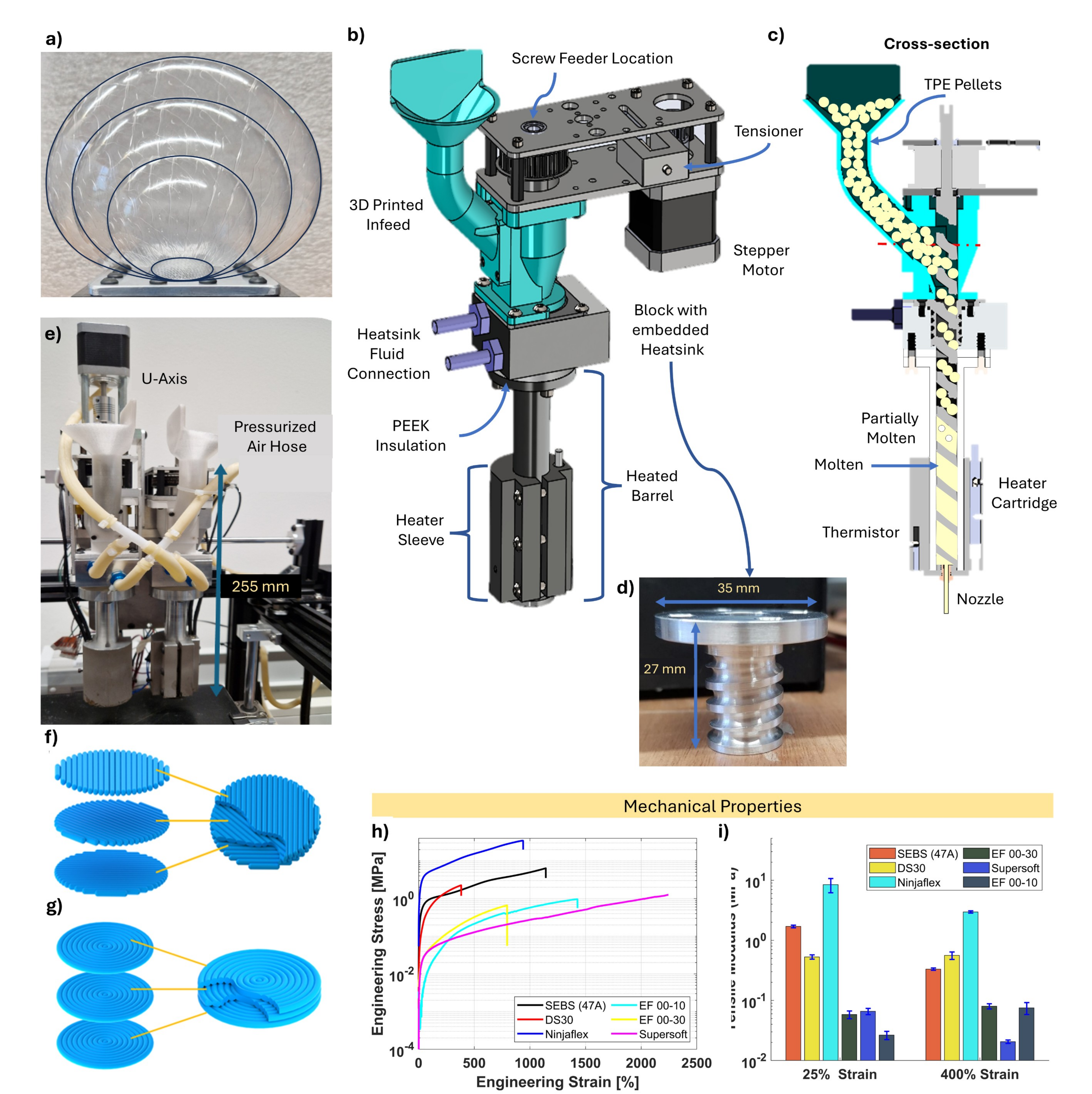}
\caption{(a) An example of a 3D-printed membrane with multiple levels of inflation indicated for clarity with the input (gauge) pressure ranging from 0 to 5 kPa, (b) the CAD model of the pellet extruder, (c) the cross-section of our pellet extruder, (d) the embedded heatsink with a helical channel, e) the pellet extruder prototype, f,g) the used infill patterns (f) lines and (g) concentric, (h) engineering stress-strain curve for a uniaxial tensile test, and (i) the moduli at low strain (25\%) and high (400\%) averaged over three samples. Abbreviations: DS30 = Dragonskin 30 and EF = Eco-flex.}
\label{fig:fig1}
\end{figure}

Thermoplastic elastomers (TPEs) are a notable class of materials that can exhibit softness and large stretchability, such as the 3D-printed membrane shown in Figure \ref{fig:fig1}(a). This combination of softness and stretching makes them suitable for soft robot applications, but also makes them challenging to 3D print with conventional filament-based MEX.

In contrast, pellet-based MEX 3D printers are an effective approach for printing very soft TPEs. However, due to the limitations and limited availability of commercial machines, we decided to develop a custom solution for this research. Inspired by industrial pellet extruders and academic designs in \cite{saari2015additive,liu2019pellet}, we developed a compact and lightweight pellet extruder, making it suitable for direct mounting on 3D printers (Figures \ref{fig:fig1}(b-e)). Our pellet extruder consists of several key components arranged from top to bottom: A drive unit, a feeder, a cooling system, a barrel with a conveying screw inside, a heater, and a printing head (nozzle). 

The drive unit, as shown in the section view (Figure \ref{fig:fig1}(c)), is connected to the conveying screw inside the barrel and controls its rotation. The feeder is a 3D-printed, funnel-shaped component designed to facilitate the loading of pellets into the machine. It guides the pellets toward the screw, ensuring they enter the screw groove properly. The rotating screw inside the barrel moves the pellets toward the heater and nozzle, where they are molten before deposition in the 3D printing process. The screw mechanism enables the extrusion and 3D printing of a wide range of thermoplastics, including challenging materials like very soft thermoplastic elastomers, by dragging the pellets instead of pushing a rod (such as in filament printers). 

Compared to pellet extruders reported in the literature, a novel feature of our machine is the inclusion of a cooling system between the barrel and feeder units. Over time, heat accumulation can cause the pellets in the feeder funnel to become hot and sticky, leading to clumping (caking). This caking disrupts the flow of pellets into the conveying screw, hindering continuous material flow during the deposition process. The cooling system prevents heat from the heated barrel from easily reaching the feeding zone, reducing caking and minimizing deposition interruptions. This system features an internal spiral heat sink (Figure \ref{fig:fig1}(d)) designed to effectively circulate the cooling fluid.

The heatsinks were connected to a pressurized air source through a flexible air hose using a barbed fitting. Two barbed fittings were used, with the right one connected to a source of pressurized air, while the left hose redirected the hot air away from the user.

Additionally, we incorporated a custom-made PEEK washer as a thermal insulator to further reduce heat transfer to the feeder area. To build this pellet extruder, we used an 80 W heater cartridge and a thermistor housed in an aluminum heater sleeve mounted around the barrel. Standard 3D printing nozzles were used throughout our printing.

For material conveyance, we employed a 12 mm wood auger drill bit (Robert Bosch Power Tools GmbH, Germany). The deep grooves of the wood auger drill provide an efficient and cost-effective solution for transporting pellets from the feeder to the heating zone. The drive unit is powered by a NEMA17 stepper motor, which is connected to the auger drill (screw) via a timing belt and reduction pulleys with a 12:40 ratio. Two copies of this pellet extruder (for multi-material printing purposes) were mounted on an Ender 5 Plus (Shenzhen Creality 3D Technology Co., Ltd., China) (Figure \ref{fig:fig1}(e)). Pressurized air was used as the cooling fluid in the cooling unit to regulate the temperature of the pellets within the feeder unit.

One common challenge when using pellet extruders for 3D printing is their nonlinear extrusion behavior. We addressed this issue by combining experimental data and manual calibration of the extrusion value (as described in the Supplementary Information) for G-Code generation. The approach for determining the other printing parameters is described in the Supplementary Information as well. An important parameter in G-Code generation is the infill pattern, which in this work was either concentric or lines (Figures \ref{fig:fig1}(f) and \ref{fig:fig1}(g)). We used custom G-Code generators based on PolySlicer \cite{willemsteinEngD} to control the printing process of soft TPEs.

To illustrate the softness of these TPEs compared to conventional materials such as silicone rubbers and filament-based MEX, we conducted uniaxial tensile tests using two TPEs (SEBS 47A and Supersoft 00-30) and other commonly used materials in soft robotics, which include flexible filament (Ninjaflex from Ninjatek, USA) and casted soft silicone rubbers (Eco-Flex 00-10, Eco-Flex 00-30, and DragonSkin 30 from Smooth-On, Inc., USA). The resulting engineering stress and strain (Figure \ref{fig:fig1}(h)) shows stresses and maximum strains across several orders of magnitude, with a comparatively steep initial slope at low strains followed by a less steep slope.

Furthermore, the engineering stress at low strain aligns with the Shore Hardness. At higher strains, stress levels correspond to Shore Hardness when categorized by material type: thermoplastics (Supersoft, SEBS, NinjaFlex) and thermosets (Eco-Flex 00-10, Eco-Flex 00-30, DragonSkin 30). Among the thermoplastics, NinjaFlex, SEBS, and Supersoft TPEs had maximum stresses of 34.6 MPa, $\leq$12 MPa, and $\leq$1.5 MPa, respectively, showing more than 20x reduction in maximum stress between the flexible filament (Shore Hardness 70A) and our softest TPE.

The uniaxial results also showed that thermosets with similar Shore Hardness, such as DragonSkin 30 and Eco-Flex 00-10, exhibited stress values comparable to the SEBS and Supersoft TPEs. However, there was no correlation between Shore Hardness and maximum strain unless the materials were grouped by thermoplastic and thermoset. Notably, the Supersoft sample did not break but reached the maximum capacity of our tensile tester (see Figure S3).

Lastly, the modulus values (averaged across three samples) are shown in Figure \ref{fig:fig1}(i). At both low (25\%) and high (400\%) strains, the modulus showed little variation. Notably, the low-strain modulus correlated with Shore Hardness, while the high-strain modulus did not, except when comparing thermosets and thermoplastics separately. An unexpected finding was the reduction in modulus for all thermoplastics at high strain, unlike silicone rubbers, which maintained or increased their stiffness. This behavior could be due to the cross-linked structure of thermosets, which enhances stiffness at higher strains. 

%%%%%%%%%%%%%%%
\subsection{Thin Inflatable Membranes}
One promising application of soft 3D-printed TPEs is inflatable membranes, which can be used in soft fluidic actuators, such as bending actuators and sucker-based devices (as demonstrated in later sections). However, 3D printing inflatable soft thermoplastic structures has been a challenge due to the limited softness of filament printers. 

\begin{figure}[h]
      \centering
      \includegraphics[width=0.98\textwidth]{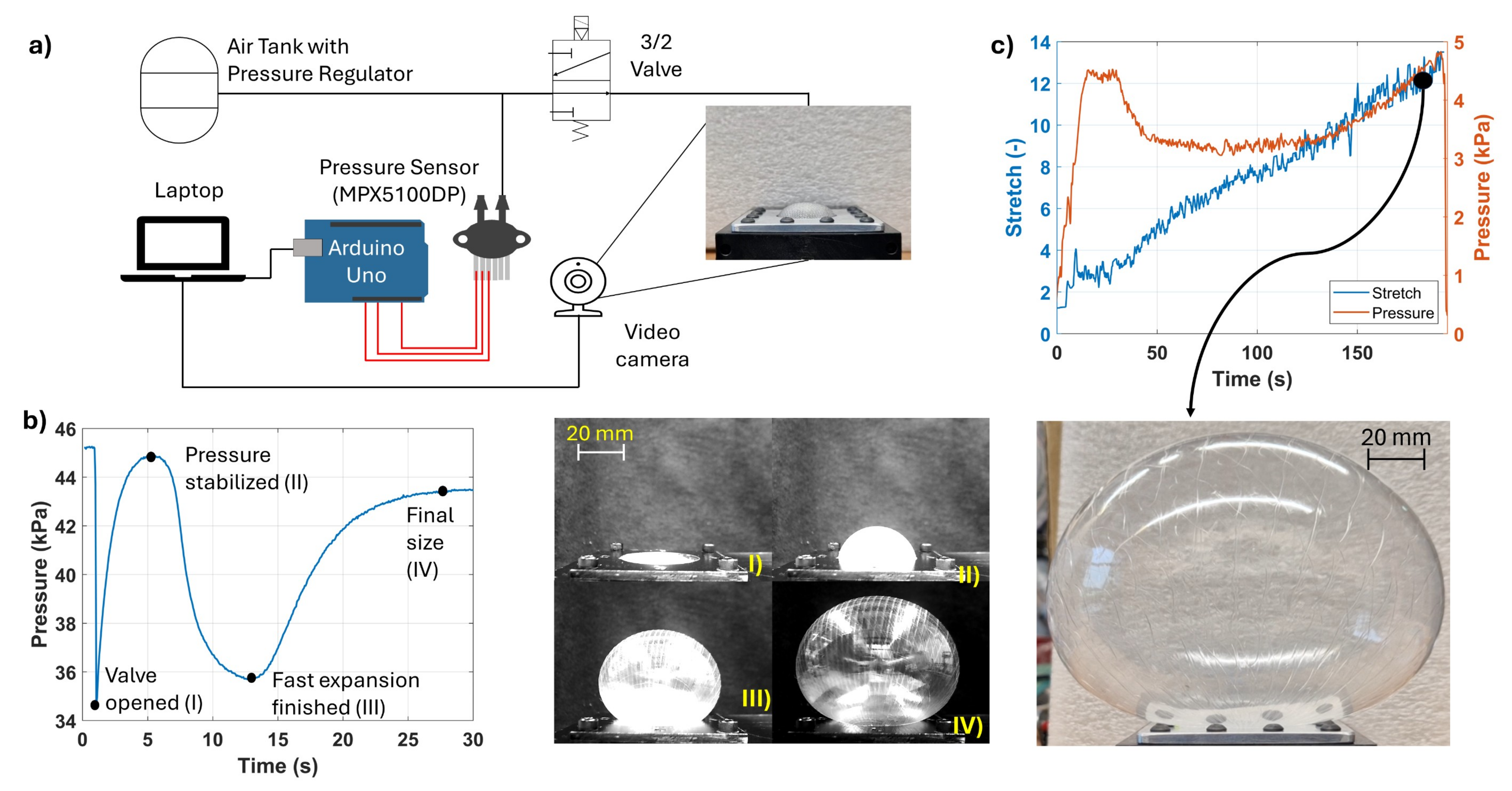}
      \caption{ a) The setup used for the inflation experiments, b,c) The expansion behavior in pictures and over time of a (b) SEBS and (c) Supersoft membranes.}
      \label{fig:Fig2}
\end{figure}

To address this challenge, we investigated 3D-printed thin soft TPE membranes by characterizing their behavior over time, analyzing the relationship between stretch versus pressure for different thicknesses and infill patterns, and performing a cyclic test. To evaluate our 3D-printed membranes, we conducted inflation tests using a custom inflation setup (Figure \ref{fig:Fig2}(a)) based on \cite{inflationpaper}. This setup incorporated a manually triggered solenoid valve with a regulated pressure source for pressure control. During inflation, the pressure and membrane deformation were recorded using a pressure sensor and a camera, respectively.

We used this setup to characterize six types of 3D-printed SEBS membranes, each with a diameter of 42 mm and printed in duplicate. They had thicknesses of 0.2, 0.6, and 1.2 mm with a layer height of 0.2 mm and either the lines or concentric infill pattern (Figures \ref{fig:fig1}(f) and \ref{fig:fig1}(g)). For single-layer membranes, we increased the extrusion rate by 50\%, which led to a membrane with a thickness of 0.3 mm. Additionally, we printed a Supersoft membrane with a thickness of 1.2 mm. Moreover, printing a three-layer membrane is shown in Supplementary Movie 1.

During the first inflation cycle of a membrane, we iteratively increased the pressure until “ballooning” occurred. This ballooning effect is a yielding-like behavior where a small increase in pressure leads to a significantly larger expansion, which also occurs in other elastomers \cite{inflationpaper}. 

When ballooning occurred, the SEBS membranes showed four distinct phases during inflation, as marked in the pressure data (Figure \ref{fig:Fig2}(b) and Supplementary Movie 2). Firstly, initial expansion; when the valve was opened (point (i)), the pressure dropped as the membrane started expanding. Next, the pressure seemed to stabilize at point (ii). However, as the membrane expanded faster than the pressure regulator could compensate, the pressure began to drop. The lowest pressure occurred at point (iii), indicating that the largest expansion was complete. Finally, at point (iv), the pressure returned to a level close to the peak at point (ii). This sequence revealed nonlinear behavior and a time delay, complicating the accurate determination of the ballooning pressure due to the fixed time frame in our experiments. After the first ballooning event, a plastic deformation was observed at zero pressure. However, our further investigation indicated the usability of these membranes through cyclic testing (later in this section) and soft robotic demonstrators.

To characterize the membrane’s mechanical response throughout the inflation process, we evaluated the membrane’s behavior under pressure before and after ballooning based on the stretch ($\lambda$):
\begin{equation}
    \lambda = \frac{L}{L_0}
\end{equation}
Where $L_0$ is the initial length (mm), and $L$ is the current length (mm). The stretch is commonly used in modeling elastomers \cite{modellinginflation} and related to the strain $\lambda=1+\epsilon$, where $\epsilon$ represents the strain.

\begin{figure}[h]
      \centering
      \includegraphics[width=0.98\textwidth]{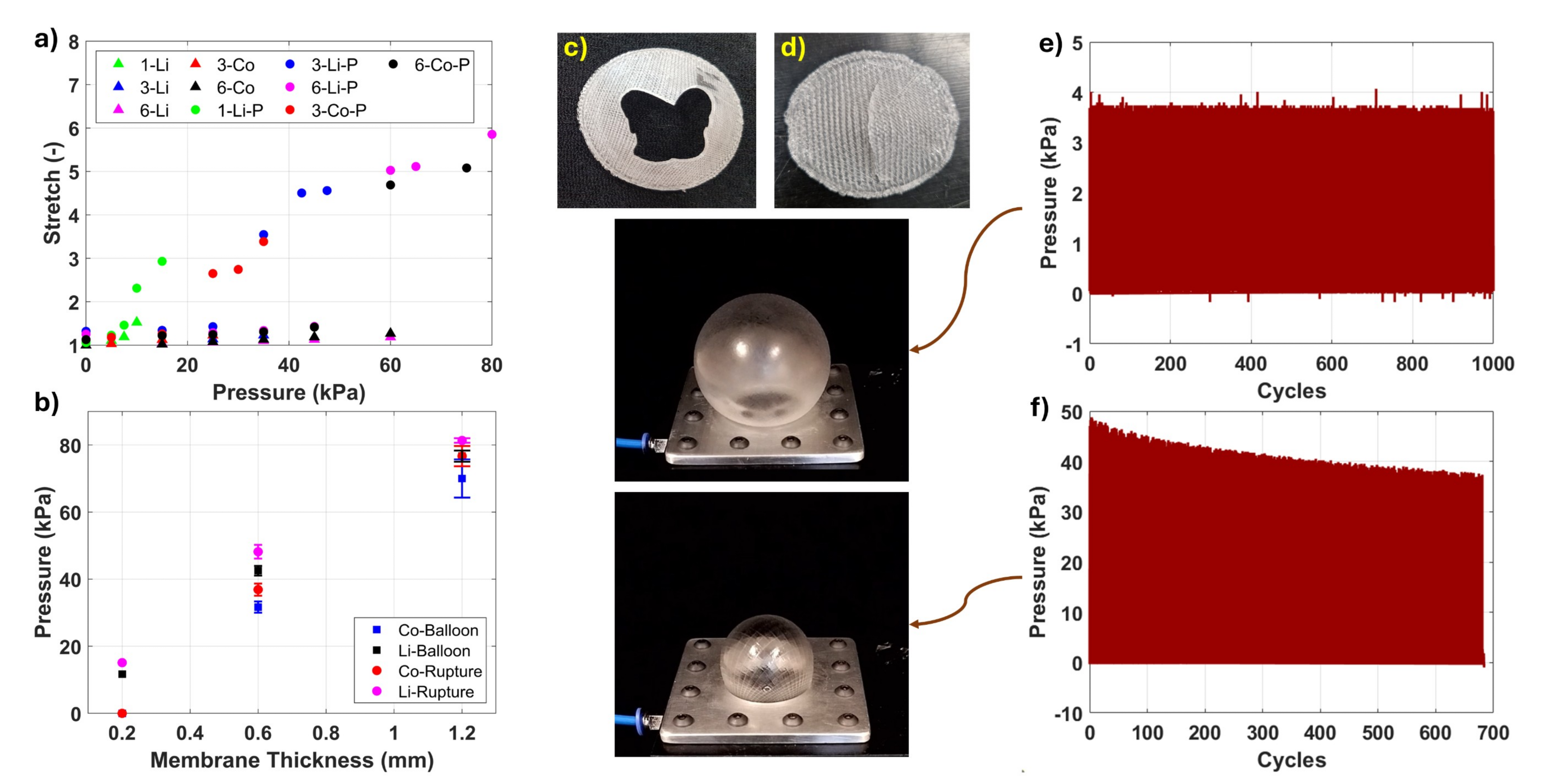}
      \caption{ a) Pressure versus stretch for the SEBS sample with the legend structured as: number of layers - (Li, Co) the lines (Li) or concentric (Co) infill and P(ost) ballooning (averaged over three samples), (b) rupture and ballooning pressures of the SEBS membranes averaged over three samples, c,d) post-rupture images of the (c) SEBS line pattern and (d) Supersoft membranes. e,f) the pressure over time for the cyclic experiment for the three-layer (e) Supersoft and (f) SEBS membranes, including an image during inflation.}
      \label{fig:Fig3}
\end{figure}

The stretch-pressure behavior of the Supersoft membrane (Figure \ref{fig:Fig2}(c), Supplementary Movie 2, and the algorithm for stretch estimation in Figure S4) was similar to the SEBS membrane. However, the maximum stretch of the Supersoft membrane (1320\% in Figure \ref{fig:Fig2}(c)) was much higher than the SEBS membrane (585\% as seen in Figure \ref{fig:Fig3}(a)). Similar to the SEBS membrane, the Supersoft membrane showed low stretch before ballooning at 4.5 kPa, followed by a sudden pressure drop, suggesting this time delay might be a general property of ballooning.

In addition to their large expansion and flexibility, an interesting feature of the membranes is their optical transparency at high stretches (Figure \ref{fig:Fig2}(c)), which could be useful for optical applications.

To characterize the stretch-pressure behavior and its relationship with printing parameters (thickness and infill pattern), we evaluated six types of SEBS membranes for a range of pressures. The resulting pressure versus stretch is shown in Figure \ref{fig:Fig3}(a) with lines (Li) and concentric (Co) infill, which does not include the single-layer concentric line, as these were not airtight. Before ballooning, the stretch increased linearly with the pressure, but at a lower rate (see also Figure S5, which zooms into the low levels of stretch). 

In addition, we observed that crossing the ballooning pressure threshold was necessary before the stretch was $>$2 (-). All membranes show a significant jump in stretch after this threshold pressure, which was followed by a slower increase in stretch. We observed that the threshold pressure before large expansion scaled approximately linearly with the membrane thickness for the line (10, 35, and 60 kPa) and concentric (25 and 60) infills. However, the maximum stretch seemed to scale with the square root of the number of layers for the line infill pattern. This correlation could be due to the thickness reduction, which scales quadratically with the stretch in inflation experiments \cite{modellinginflation}. 

Next to the pressure-stretch relationship, we recorded the rupture and ballooning pressure of the SEBS membrane (Figure \ref{fig:Fig3}(b)). These results showed that the line pattern consistently required higher pressures than the concentric pattern, suggesting that the printing pattern impacts overall performance. This localized effect of the infill pattern can be used to program the deformation, which is another advantageous capability of 3D printing. In addition, there seems to be a margin between rupture and ballooning pressures.

Images of ruptured membranes are shown in Figures \ref{fig:Fig3}(c,d) for the SEBS (line infill) and Supersoft membranes, respectively (concentric infill is provided in Figure S7). The dome shape of the SEBS membrane is due to the plastic deformation. The butterfly pattern after rupture was unexpected, as we expected a failure at the interface of the infill pattern. In contrast, the Supersoft membrane did fail at such an interface. However, it remains unclear whether this failure was due to a printing defect, the larger thickness reduction, or a combination of both.

Lastly, to investigate the usefulness of the TPE membranes for soft actuators, we performed a cyclic experiment to investigate their behavior under repeated loading and unloading. Specifically, we investigated it for two Supersoft and SEBS membranes (both were printed with three and six layers). The two three-layer membranes are shown in Figures \ref{fig:Fig3}(e) and \ref{fig:Fig3}(f) for the Supersoft and SEBS membranes, respectively. In contrast, both six-layer membranes are shown in the Supplementary Materials, and all four membranes in Supplementary Movie 3.

The pressure magnitude in the cyclic experiments indicated that both Supersoft membranes remained airtight over the 1,000 inflation/deflation cycles. The pressure behavior stayed consistent over all the cycles. In contrast, the SEBS membranes failed after 681/740 cycles for the three-/six-layer membranes. A noticeable difference between the Supersoft and SEBS membranes is the pressure magnitude over time. The Supersoft membrane’s pressure stayed the same over all cycles. In contrast, the SEBS membrane’s pressure decreased exponentially over time until failure, which could be due to some softening effects or an increase in the maximum expansion at the end of the cycle. 

Another noticeable difference was their optical appearance during inflation, with the Supersoft appearing uniform, while the SEBS membranes showed the line infill.

%%%%%%%%%%%%%%%
\subsection{Ballooning Bending Actuator}
The softness of the TPEs enables large deformations at low pressures, making them useful for soft fluidic actuators. Furthermore, soft fluidic actuators require hollow structures, which are possible through 3D printing. Making 3D-printed TPEs a promising approach for fabricating soft fluidic actuators.

\begin{figure}[h]
      \centering
      \includegraphics[width=0.98\textwidth]{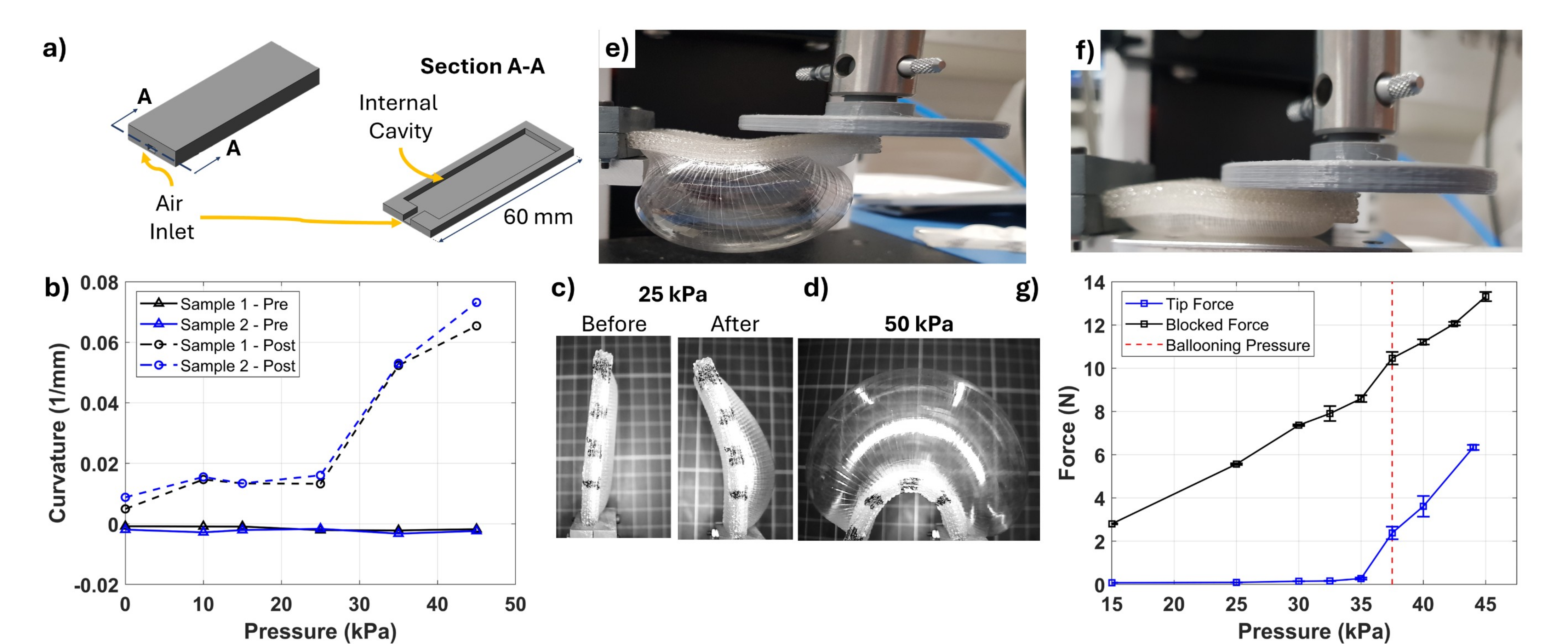}
      \caption{a) The ballooning bending actuator design, b-d) unconstrained bending behavior (b) curvature versus pressure, c,d) bending (c) before and after ballooning, and (d) maximum curvature, (e,f) experimental setups for the (e) blocked and (f) tip force, and (g) results of both force experiments averaged over three repetitions.}
      \label{fig:Fig4}
\end{figure}

To demonstrate the feasibility of 3D printed TPEs for this purpose, we fabricated a soft bending actuator 60x20x5 mm$^3$ (length x width x height with others in Figure S8) using a single three-layer membrane (Figure \ref{fig:Fig4}(a)). The design was intentionally chosen to be sub-optimal, to emphasize that substantial bending deformations are achieved through material softness and extension ratio rather than relying on geometrical features typically used with flexible filaments. 

We used a SEBS membrane to fabricate the bending actuator, which is expected to produce higher output forces. Additionally, since it is relatively hard to achieve large expansions for this TPE compared to our Supersoft TPE, we chose SEBS to show that even our less soft TPE can still undergo sufficient expansion while providing higher output forces in the actuator.

We characterized the bending actuators in terms of force and curvature. In these experiments, we reused the inflation setup for pressure control. The curvature experiment results (Figures \ref{fig:Fig4}(b)-(d)) show a significant difference between the curvature before and after ballooning pressure. Before ballooning, only minor curvature was present (Figure \ref{fig:Fig4}(c)) with a linear increase of the curvature (Figure \ref{fig:Fig4}(b)) over the range of examined pressures. After ballooning, the actuator reached a significantly larger curvature (Figure \ref{fig:Fig4}(d) and Supplementary Movie 4) with two distinct slopes separated by the ballooning pressure. The maximum deflection during ballooning (Figure \ref{fig:Fig4}(d)) was nearly 180 degrees at 50 kPa. In addition, the direction of the curvature changed after ballooning from the right to the left side, as seen in Figure \ref{fig:Fig4}(c). 

Subsequently, the output force was characterized for one of the actuators (after relaxing overnight) in both a blocked \cite{yap2016high} and tip force \cite{khondoker2019direct} scenario (Figures \ref{fig:Fig4}(e) and \ref{fig:Fig4}(f)). Similar to the curvature, the tip force versus pressure (Figure \ref{fig:Fig4}(g)) showed significant differences before and after ballooning. With a much steeper slope of the force after ballooning compared to before. This significantly steeper slope could be due to the large increase in deformation and surface area that happens after ballooning.

In the blocking force experiment, the slope of the force around the ballooning pressure is different. However, the slope before and after ballooning does seem similar, unlike in the curvature and tip force experiments. In addition, due to the mechanical constraint, the membrane did not balloon in size (Figure \ref{fig:Fig4}(e)). Due to the lack of ballooning, the membrane thickness did not reduce as much, which led to higher maximum pressures than observed when the membrane was unconstrained. The output force of this actuator had a maximum of 6.34 N and 13.33 N for the tip and blocked force, respectively. These forces are between 113 and 238 times its weight (5.7 grams). 

%%%%%%%%%%%%%%%%%%%%%%%%%%%%%%%%%%%%%%%%%%%%%%%%%%%%%%%%%%%%%%%%%%%%%%%%%%%%%%%%%%%%%%%%%%%%%%%%%%%%%%%%%%%%%%%%%%%%%%%%%%%%%%%%%%%%%%%%%%%%%%%%%%%%%%%%%%%
%%%%%%%%%%%%%%%
\subsection{Membraned Soft Sucker with Integrated Fluidic Sensing}

The capability of 3D printing airtight membranes enabled us to fabricate membraned suckers inspired by the design of sea urchin-like suckers \cite{sadeghi2012design}. These suckers use the membrane deformation to generate a vacuum and adhesive force, the softness to seal themselves to the substrate, and protection from contaminants \cite{mazzolai2019octopus}. Moreover, in this work, we present a novel approach that uses these membranes as a tactile sensor by detecting the increase in internal pressure when an external load is applied. This approach makes the membraned suckers sensorized, allowing them to detect objects before activating adhesion. The integration of actuation and sensing in a single structure reduces the need for additional components, simplifies system design, and enables direct force measurement at the point of contact, enhancing the responsiveness and functionality of the soft actuator.

\begin{figure}
      \centering
      \includegraphics[width=0.98\textwidth]{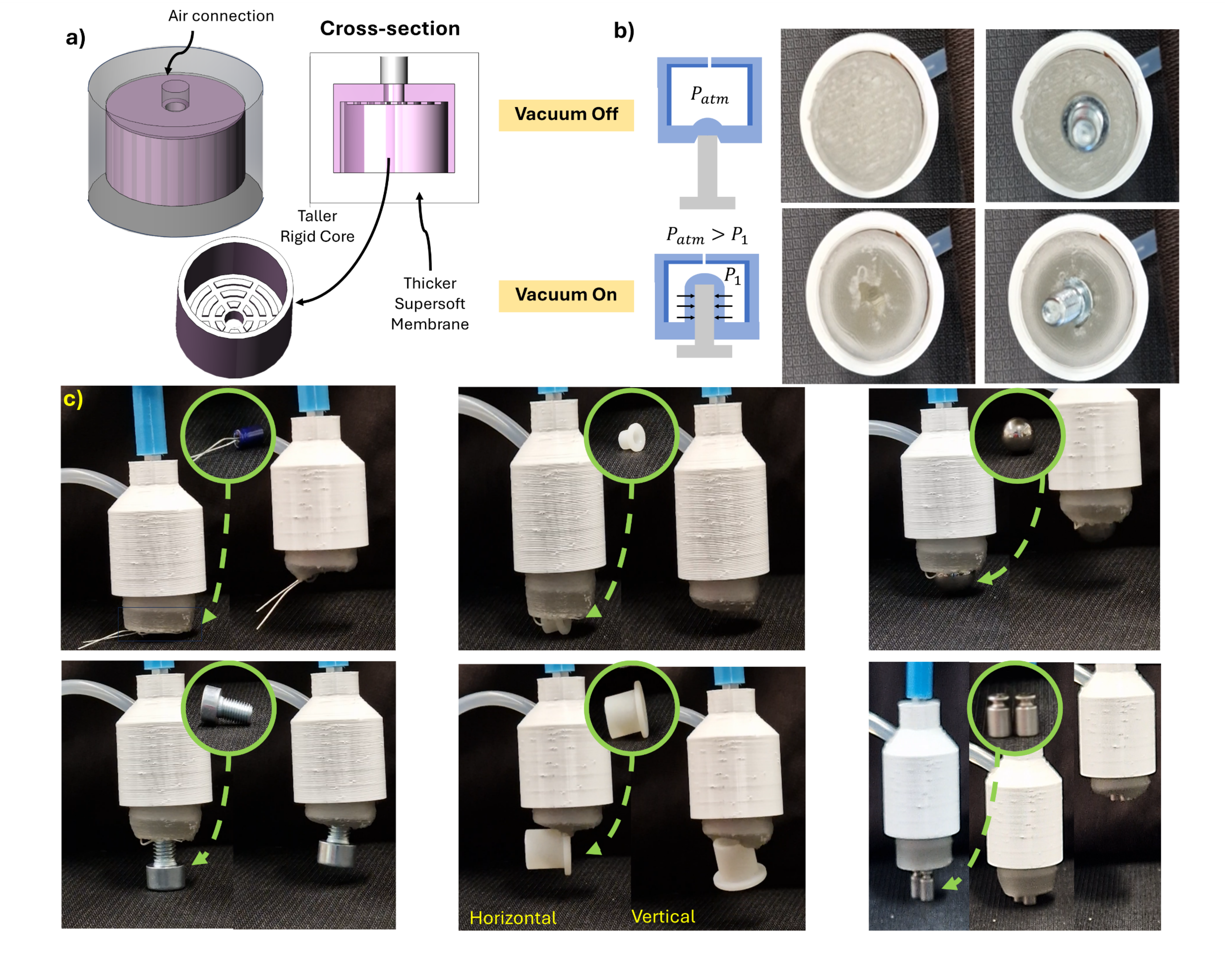}
      \caption{a) The sensorized sucker design, (b,c) behavior of sucker under vacuum (b) conceptually and (c) with the realized sucker, (d,e) adhesion force (averaged over three measurements) with different (d) substrates (with and without gel applied to the soft foot) and (e) pre-loads, (f) inflating the sucker to enable tactile sensing, (g) the pressure sensing behavior over time, (h) pressure versus force plot of the sensor for multiple cycles, (i-k) grasping of (i) flexible sheet, (j) mug, and (k) styrofoam with S, T, and A indicating the Start, Touch, and Adhesion of the suckers. The pressure is positive during sensing (start (S) to touch (T)) and below ambient ($\approx$100 kPa) during adhesion (A).}
      \label{fig:Fig5}
\end{figure}

The membrane-based sucker (Figure \ref{fig:Fig5}(a) with dimensions in Figure S9) benefits from the TPE's softness for adaptation, adhesion, and sensing. Firstly, the soft foot of the membrane can adapt to a surface using the soft TPE’s adaptability, enabling sealing by mechanical compliance. Secondly, the adhesion force increases as a softer (i.e., less stiff) membrane will deform more under the same input vacuum (Figure \ref{fig:Fig5}(b)). This larger deformation increases internal volume, which correlates with a higher vacuum \cite{sadeghi2012design}. Thirdly, sensing profits from the softness by being inherently more compliant, which reduces the force needed for barometric changes.

Due to the aforementioned benefits of a softer membrane in these suckers, we used our Supersoft TPE. Furthermore, following the design presented in \cite{sadeghi2012design}, we placed a rigid core inside the soft sucker cavity to prevent undesired deformations and maximize membrane deformation. Additionally, this rigid core supports the soft structure for better force transmission to the surface \cite{sadeghi2012design}. 

For the fabrication of this sucker, we combined filament and pellet printing (as shown in Figure S10). The resulting sucker had an external diameter of 22 mm, a membrane thickness of 3 mm with a 45-degree chamfer, and an internal cavity with 2 mm-thick walls. The rigid core is a cup with a diameter of 18 mm, 2 mm walls, and a height of 13 mm printed out of PLA. As seen in Figure \ref{fig:Fig5}(c), this rigid core squished the membrane for sealing when a vacuum is applied.
 
Furthermore, the sea urchin-inspired sucker in \cite{sadeghi2012design} demonstrated enhanced sealing on rough surfaces by using a viscous gel as a replacement for mucus in biological tube feet. In this approach, the gel fills potential gaps/surface roughness to improve sealing even for samples with high roughness. To investigate the effect of such a gel on our sucker's adhesion, we added a muscle gel (Spirosan, The Netherlands) as an artificial mucus on the soft foot (Figure \ref{fig:Fig5}(a)). We observed a significant increase in adhesion force (Figure \ref{fig:Fig5}(d)) at a low pre-load from below 1 N to $>$10 N. As a comparison, we also repeated this experiment for an increasing pre-load (Figure \ref{fig:Fig5}(e)). As the pre-load increased, the adhesion forces became similar (around 12 N) to those with the gel, suggesting that the Supersoft TPE can adapt through its softness and self-seal.

To enable sensing, we inflated the membrane (Figure \ref{fig:Fig5}(f)) to achieve a concave shape. This transformation allows the sucker to function like a finger with tactile capabilities to detect objects before attachment. In our setup, the concave shape was achieved by having a pump connected to a partially open flow resistor (the fluidic circuit is shown in Figure S11). We characterized the pressure sensing by pushing cylindrical probes of different sizes against the sucker (3 mm, 6 mm, and 12 mm) with the data for the 12 mm probe shown in Figures \ref{fig:Fig5}(g,h) and the others in Figure S12. We observed that the fluidic sensor is sensitive to low forces ($<$0.25 N), with the pressure increasing with the force (as expected). In addition, a low hysteresis was present, which is beneficial for consistent force sensing. The signal recorded from small objects, combined with the sucker's movement, can help predict object size and prevent unnecessary effort in attempting to grasp objects that are too small to adhere to. However, this capability requires further research, which is beyond the scope of this work.

Lastly, we tested the sucker’s tactile sensing and adhesion in three grasping tasks (Figures \ref{fig:Fig5}(i-k) and Supplementary Movie 5). The pressure is above ambient between start (S) and touch (T) due to inflation of the sucker for sensing, and below ambient (the start value) during adhesion (A). During touch, the pressure increased, illustrating the force sensing capabilities. In these scenarios, the sensorized sucker detected pressure changes and adhered to the objects, demonstrating that the TPE membranes can be used for 3D-printed sensorized suckers that leverage their softness for sensing and self-sealing.

%%%%%%%%%%%%%%%
\subsection{Membrane-based Gripper}

\begin{figure}
    \centering
      \includegraphics[width=0.98\textwidth]{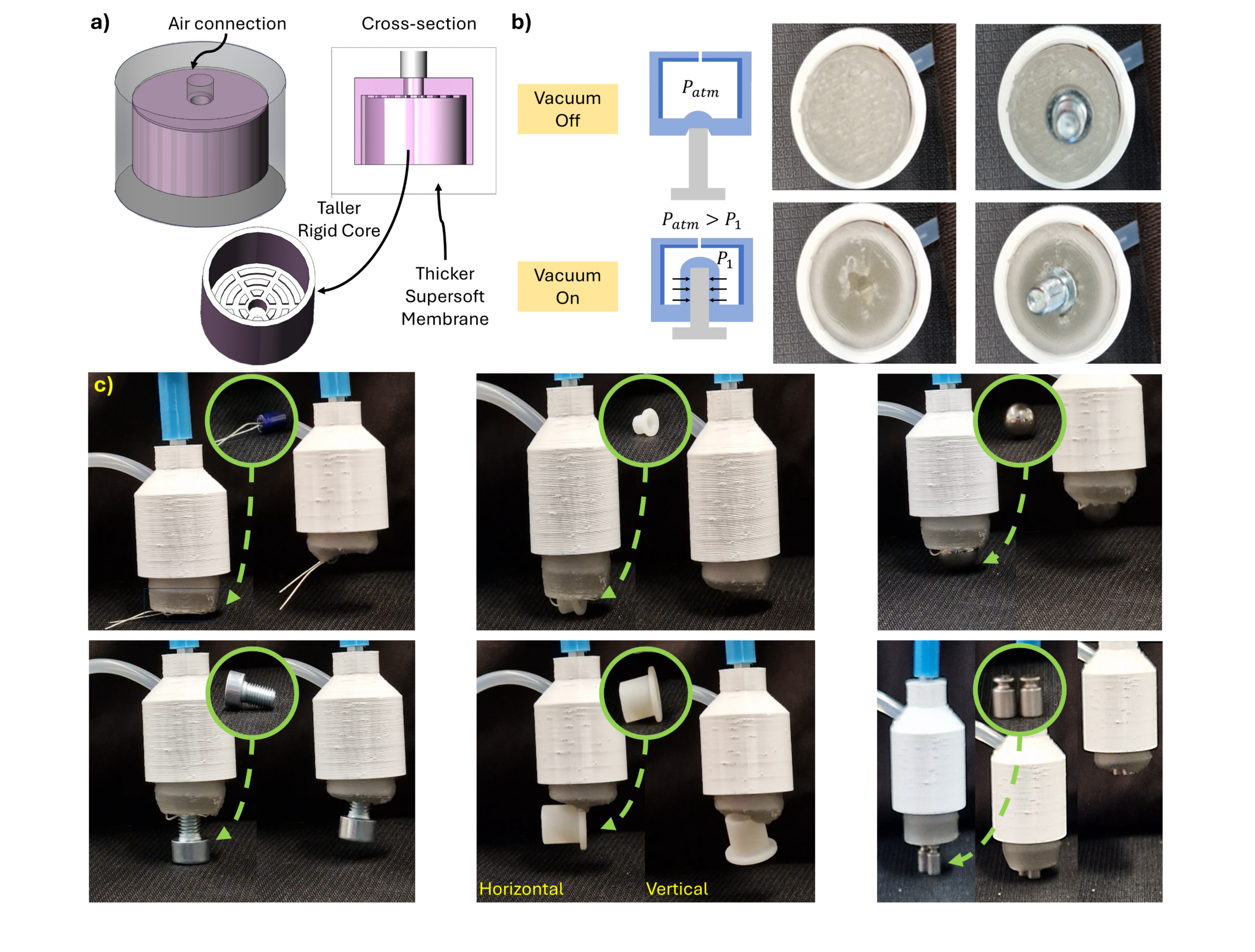}
      \caption{a,b) The membrane-based gripper (a) design and (b) actuation behavior including working principle, and (c) using the membrane-based gripper to grasp objects with a wide variety of shapes and sizes both bigger and smaller than the membrane itself, which were a capacitor, a small (compared to gripper size) nylon bearing, a large nylon bearing (both horizontally and vertically), a ball bearing, a bolt, and two small weights at once.}
      \label{fig:Fig6}
\end{figure}

During our experiments with different membrane sizes for sucker adhesion, we observed a notable change in the membrane's deformation behavior. Specifically, using a thicker membrane (8 mm) and a longer rigid core (Figure \ref{fig:Fig6}(a) with dimensions in Figure S9) in the sea urchin-inspired sucker design led to the membrane no longer sealing effectively against substrates. Instead, it displayed a grasping-like deformation capable of grasping various objects (Figure \ref{fig:Fig6}(b)). This actuation method, to the best of the authors’ knowledge, has not been discussed in the literature. Notably, it allows for the gripping of objects smaller than the membrane’s diameter, unlike membraned suckers that can only grasp objects bigger than their membranes, as the suckers require sealing for adhesion. This other grasping method profits from the adaptability and high friction of the Supersoft TPE, as it needs to conform around the object for grasping.

Through the TPE’s mechanical compliance, the gripper could grasp a wide variety of objects (Figure \ref{fig:Fig6}(c) and Supplementary Movie 6), which were bigger and smaller than the membrane. These objects included a capacitor, a bolt, a weight, and a ball bearing. Additionally, it could lift objects with different orientations, such as the larger (compared to gripper size) nylon bearing. In these scenarios, the membrane displayed two grasping modes, depending on the object's size relative to the membrane's diameter. When the object is larger, the membrane jammed a part of the object's surface. In contrast, smaller objects are swallowed and moved inside the rigid core. The latter could be useful to protect an object during transport. Lastly, through the TPE’s adaptability, we could grasp multiple objects simultaneously. 

We observed that the gripper was limited to a minimum size as the membrane collapsed to a circular shape of approximately 4 mm (see Figure \ref{fig:Fig6}(b)). Nevertheless, the gripper demonstrated another use case of membranes that benefit from the soft 3D-printed TPE's ability to conform to an object, providing another type of 3D-printable soft gripper. However, further research is needed to understand its behavior and improve grasping.

%%%%%%%%%%%%%%%%%%%%%%%%%%%%%%%%%%%%%%%%%%%%%%%%%%%%%%%%%%%%%%%%%%%%%%%%%%%%%%%%%%%%%%%%%%%%%%%%%%%%%%%%%%%%%%%%%%%%%%%%%%%%%%%%%%%%%%%%%%%%%%%%%%%%%%%%%%%%%%%
\section{Discussion}

The 3D printing of hyperelastic materials is important for increasing the complexity of soft and hybrid robots. TPEs are an interesting class of materials for this purpose as they combine softness with the benefits of thermoplastics, such as thermal re-processing and rapid solidification. 

Besides the advantages of thermoplastics, the uniaxial results show that the uniaxial engineering stress of the Supersoft TPE is similar to that of the commonly used soft Eco-flex 00-10 silicone rubber. Furthermore, we could not break our softest TPE sample, suggesting a high maximum elongation, which was also observed in \cite{curmi2025screw}. Although we cannot verify whether our sample could also achieve their reported 3365\% strain, our sample reached a strain similar to their Z-direction sample (2328\%). However, recent work on polypropylene elastomers \cite{bayati20243d} suggests even higher elongation can be reached with pellet-based printing (4048\%), even though they used a much higher Shore Hardness material (64A).

Furthermore, our findings in the uniaxial experiment suggest that Shore Hardness is only a reliable indicator of the modulus at low strains. For instance, the 00-30 TPE and 47A SEBS were less stiff at higher strains than silicone rubbers with lower Shore Hardness, such as Eco-flex 00-10 and Dragonskin 30A. Additionally, the difference in the change of modulus for thermoplastics and thermosets at higher strains was notable and could be exploited in future designs for sequential activation. 

One limitation of our work is that we used a single strain rate, which can affect stress measurements due to viscoelastic behavior. Although these effects may favor certain polymers, it is not expected that they would significantly affect the mechanical similarity. In a future work, other characterization methods such as DMA and DSC can be performed to compare the (thermo-)mechanical behavior in-depth of the Supersoft/SEBS TPE and its silicone rubber counterparts.

Subsequently, we used the soft TPEs to print thin airtight membranes. Notably, we could print airtight single-layer membranes. We expect that it is feasible to print smaller airtight strips or membranes as well, as small imperfections would have created leaks on the larger surface area. We hypothesize that this ability to fabricate thin and airtight structures is possible due to the lower viscosity of these soft TPEs, which enables the material to flow into gaps when extruding. As suggested by the uniform expansion seen during the cyclic tests of the Supersoft TPE. 

In the cyclic tests, the Supersoft TPE membranes survived 1,000 inflation/deflation cycles for both the three- and six-layer membranes, suggesting that these Supersoft membranes can be useful for soft actuators. However, the SEBS membranes survived significantly fewer cycles. More research is required to understand the failure mechanisms, as our results are inconclusive regarding the underlying reason for the difference. One possible explanation could be the printing imperfection, which was more pronounced in the SEBS membrane, as the infill pattern is visible during expansion. The stress concentration at these imperfections can negatively impact the lifespan. However, other possible causes include inflation time, pressure, maximum expansion during inflation, its softness, or perhaps a combination. A better understanding of these parameters is needed to optimally use these 3D-printed membranes in soft robots.

In addition, we used the SEBS and Supersoft TPE membranes for three demonstrators: the bending actuator, a sea urchin-like sucker with tactile sensing, and a membrane-based gripper. These demonstrators showed that the 3D-printed membranes were, besides being airtight, capable of large deformations, self-sealing on rough substrates, sensing, and grasping. 

Moreover, these demonstrators benefited from the use of TPEs with 3D printing by enabling the direct fabrication of soft hollow structures and the integration of rigid 3D-printed cores. This capability eliminated the need for complex molding and multi-part assembly, demonstrating the mutual advantages of combining soft TPEs with pellet-based MEX.  

We also benefited from the large expansion of the TPE membranes (585\% and 1320\% for SEBS and Supersoft membranes, respectively) to realize the bending actuator's large deformation. Our bending actuator used 50 kPa for 180-degree bending without any geometrical tricks, while 150 \cite{yap2016high} kPa and 130 \cite{khondoker2019direct} kPa were necessary for Ninjaflex and SEBS PneuNet actuators, respectively. Implying that the PneuNet required more pressure than the membrane. Especially as the SEBS in \cite{khondoker2019direct} was the same material with 1 and 1.5 mm walls (chambers and lower portion), which is comparable to our membrane. However, this analysis might be biased as the PneuNet has more geometrical conditions, which makes absolute comparisons of thicknesses difficult. Furthermore, the PneuNets reached higher deformation (up to 270 degrees and 360 degrees for the Ninjaflex PneuNet) and could withstand higher pressure (400 kPa), which are favorable properties in some applications. 

Nevertheless, the ballooning membranes simplified the design yet provided a reasonable curvature (180 degrees) without any geometrical tricks. Comparing the output force of the Ninjaflex PneuNet with ours, our actuator had a significantly lower output force (70.4 N versus 13.33 N) but a similar force per kPa (0.23 and 0.27 N/kPa), implying a similar conversion to useful force. However, the SEBS PneuNet had a tip force at 400 kPa of 1.4 N while ours required 50 kPa to reach a tip force of 6.34 N, implying the ballooning actuator is more efficient in converting to a force than the PneuNet. Thus, even though our design is simpler, the combined softness and ballooning can outperform PneuNet for force. However, it is unclear how optimization would affect both the PneuNet and the ballooning actuator. For instance, increasing the membrane thickness increases the maximum pressure and could increase its deformation/force. In addition, our membranes showed plastic deformations after ballooning, which could be unfavorable and avoided through a PneuNet design.

Furthermore, our bending actuators (and membranes) show that the stretch-pressure relationship changes above the thickness-dependent ballooning pressure. This behavioral change could be used to program a deformation by locally changing thickness and/or pressure regulation. This behavior is different from other strategies, such as \cite{kim2019bioinspired}, which rely on a combination of material properties, geometry, or origami. Therefore, ballooning provides designers with another tool for programming deformation behavior.

Besides the bending actuator, the membrane-based sucker and gripper also showed the potential of the soft membranes for temporary adhesion and grasping purposes. Their inherent adaptability was essential for the membrane-based gripper to envelop the object and create a large friction for secure handling. For membrane-based suckers, this adaptability enables self-sealing, significantly enhancing adhesion forces. Our results suggest that a higher pre-load was necessary to achieve the same adhesion force as with mucus. The increase in force due to the presence of the mucus was a factor of more than ten times at the low pre-load, which was higher than the highest difference reported in \cite{sadeghi2012innovative}. This effect could be due to the higher surface roughness of our 3D-printed suckers, which require a higher initial pre-load or gel to fill the gaps before a proper seal can be formed.

Additionally, the airtightness of the membranes opened up the possibilities for integrating barometric soft sensing into suckers/grippers for tactile sensing. This approach could detect low forces and had a low hysteresis, offering a significant advantage over conventional suction cups, which only provide binary vacuum feedback. Future work could explore how the pressure profile can be used to determine optimal grasping locations or assess an object's softness. The latter was possible for a different suction cup type with sensing in \cite{zou2024retrofit}. 

Zooming in on the inflation behavior of the membranes, we observed a slightly larger expansion of the line infill pattern. This larger expansion can be exploited to program the deformation of the membrane , benefiting the ability of 3D printing to locally pattern the structure. Future research could explore incorporating intentional design variations to realize balloons with pre-programmed deformation behavior for grasping or other applications. Introducing complicated patterns can lead to unique deformation behaviors, such as in \cite{pikul2017stretchable}, which could be interesting to combine with ballooning. 

The observed ballooning effect is interesting, as it reduces the need for higher pressures, simplifies the design, and delivers performance comparable to a SEBS PneuNet (as discussed earlier). However, ballooning is often considered a problem leading to puncture and undesired motions \cite{gariya2023experimental,he2023rubber}. This issue is further complicated by the proximity of the ballooning and rupture pressure and the presence of undesirable plastic deformations. Therefore, more research is needed to understand the ballooning effect and its relation to mechanical properties, geometry, and time. For instance, how does the ballooning pressure vary with thickness, elastomer properties, and cyclic behavior? Our cyclic testing results provide an initial indicator that the Supersoft TPE membrane can survive at least 1,000 cycles. However, more cycles and understanding the failure mechanisms, such as those seen in the SEBS membrane, are still needed (as discussed earlier).

In addition, a time delay was present (Figure \ref{fig:Fig2}(a)), making the analysis more complicated. Therefore, research is needed to characterize the effect of pressure magnitude, pressure history, and time spent under pressure. In addition, the threshold-like behavior makes it suited for valves, while its transparency changes are interesting for optical applications. Making ballooning challenging yet potentially rewarding. In addition, it is important to recognize that the slow pressure source likely made the time delay more noticeable. Nevertheless, understanding the time delays and nonlinear pressure changes is necessary for accurately controlling the pressure to prevent bursting the membrane and understanding the behavior of the soft TPE material. Future work should explore how different flow rates affect the membrane, as both pressure and flow rate can be useful tools to control the membrane/soft actuator.

Zooming out to the printing process itself, we showed a pellet extruder design featuring a novel cooling system with a passive thermal insulator and a single pressurized fluid. Through its simplicity, it can be easily integrated into other screw extruders with a cylindrical barrel. However, one challenge we encountered was the need to adjust extrusion parameters for different movements and starting/stopping the extrusion. This step took considerable time and is a point of improvement in future work. One possible solution is to separate the melting and extrusion by using two separate systems, such as the pneumatic system in \cite{mirasadi20243d,bayati20243d}. Separating the melting and flow control could allow for more fine-tuned control of the extruded amount.

Taking a step further back, how does pellet-based MEX fit into the 3D printing of soft robots using thermoplastic (elastomers)? To give an indication, we compared multiple MEX processes and selective laser sintering (SLS) in Table \ref{tbl:Compare}. The pellet extruders seem to fit a niche between ink-based and filament-based methods. Having advantages in terms of softness and simplicity of the machine at the cost of more complicated extrusion control, but with less material range than ink-based methods. However, it does not rely on curing, which, if not sufficient, can make the structure unstable \cite{dong2024application}. Although fast-curing polymers can minimize this issue \cite{yamagishi2024direct}.
\begin{center}
\begin{table}[]
    \centering
    \begin{tabular}{ C{8em} C{6em} C{6em} C{6em} C{6em}} 
      \hline
      \textbf{Technology} & \textbf{Filament-Based MEX} &  \textbf{Pellet-Based MEX} &  \textbf{Ink-Based MEX} &  \textbf{SLS}\\ 
      \hline
      Base Material & Filament & Pellets & Inks & Powder \\
      Softness (Shore Hardness) & $>$60A \cite{blanco2024manufacturing} & 00-30 (this work) & 00-10 \cite{yamagishi2024direct} & 45A \cite{liu2025advances}  \\
      Solidification & Thermal & Thermal & Curing & Thermal \\
      Extrusion Method  & Pushing & Screw/ Pneumatic\cite{bayati20243d} & Air pressure/ Piston/ Screw \cite{liu2025advances} & n/a \\
      Extrusion Control & Simple & Nonlinear & Simple & n/a \\
      Multi-material \cite{dong2024application,blanco2024manufacturing,liu2025advances} & Yes & Yes &  Yes & Difficult \\
      System Complexity \cite{dong2024application,liu2025advances} & Low & Low & Low & High \\
      Cost & Mid & Low \cite{goh2024large} & Varies & High \cite{dong2024application}\\
      Large Print Scaling \cite{goh2024large} & ? & Possible &? & Possible \\ 
      Material Range \cite{dong2024application} & + & ++ & +++ & +\\
      \hline
    \end{tabular}
    \caption{Comparison of 3D printing technologies for soft robot applications. System complexity involves: material handling, process control, and post-processing. Due to the wide range of inks, the material cost is not evaluated, while filament is compared to pellet cost. Abbreviation: MEX = material extrusion, SLS = selective laser sintering. A larger number of + indicates a relatively larger range in materials, whereas the "?" indicates that it is unclear whether large print scaling would be feasible.}
    \label{tbl:Compare}
\end{table}
\end{center}

The pellet-based MEX enables the printing of softer materials at a low process complexity (as pellets are readily available). Notably, pellet-based MEX can scale to much larger prints \cite{goh2024large}, which combined with the low material costs, can be beneficial for larger prints that are (partly) soft. In addition, unlike selective laser sintering (SLS), pellet extruders combine softness, low process complexity (less post-processing and powder handling), and straightforward multi-material integration. The differences between the different printing methods seem complementary and solve a weakness. Therefore, perhaps instead of thinking of the other methods as competitor technologies, it would be more beneficial to combine their relative strengths, such as in the hybrid casting and printing methods \cite{rossing2020bonding,goshtasbi2025bio}.

This complementary nature can be exemplified by examining the broader picture of filament and pellet-based MEX processes; the range of moduli available by combining pellet and filament extruders is very significant. More specifically, our printed TPEs reached around 100 kPa, while carbon fiber-reinforced nylon PolyMide PA12-CF (Polymaker, China) can reach 3.34 GPa. Although four orders of magnitude is significant, it is not near the human body that spans Pa to GPa \cite{guimaraes2020stiffness}. Mimicking this range would make it easier to mimic biological systems. Extending the range could involve adding gels and continuous fiber-reinforced thermoplastics to provide a single 3D printing method with multiple (smart) materials and a large stiffness range. We anticipate that adding flexible fibers to restrict expansion could increase the maximum pressure. This expectation is based on our blocked force experiment, where limiting the expansion of the membrane resulted in higher maximum pressure. The 3D printing of such actuators still profits from the soft TPEs for soft airtight balloons, as lower balloon stiffness reduces the required pressure to achieve the same output force in the actuator.

%%%%%%%%%%%%%%%%%%%%%%%%%%%%%%%%%%%%%%%%%%%%%%%%%%%%%%%%%%%%%%%%%%%%%%%%%%%%%%%%%%%%%%%%%%%%%%%%%%
\section{Conclusion}
In this work, we explored the potential of soft (Shore Hardness $<$50A) and very soft (Shore Hardness 00) TPEs to fabricate soft structures with a stiffness similar to soft silicone rubbers using pellet extruders. We utilized the softness and airtightness of the 3D-printed TPE structures to print membranes that can expand significantly, which we used to realize soft (sensorized) actuators. 

These soft demonstrators combined the mechanical compliance of soft TPEs with the versatility of 3D printing to realize hollow structures and combine soft and hard materials. Although more work is needed to fully utilize these TPEs and/or their membranes (as discussed earlier), our results show promise for using 3D-printed (very) soft TPEs in soft robots.

By utilizing these (very) soft TPEs we could print mechanically compliant structures without geometrical tricks. Future work could extend our results by investigating methods to combine TPEs with standard 3D printing filaments such as PLA and Ninjaflex. Such a study could focus on the bonding of soft TPEs to rigid/flexible materials. Interfacing such mechanically different materials could, for instance, use mechanical interlocking, such as in \cite{grignaffini2024new}. Developing appropriate bonding strategies is crucial to realize structures from very soft (Shore Hardness 00) to rigid materials in a single process. Such a strategy might consist of pellet-based and filament MEX, but perhaps also ink-based MEX, to widen the available material range at the cost of increased process complexity (i.e., how to bond filament/pellet with inks).

Besides bonding strategies, a better understanding is needed to simplify the extrusion control, as our iterative experimental approach is time-consuming. A possible solution is to add a system between the nozzle and the pellet extruder. This intermediate system should stabilize and control the volumetric flow without the pellet extruder's complex flow behavior. Lastly, the sucker results with the gel suggest that lowering the surface roughness of the printed structure is important for higher adhesion forces (in line with \cite{sadeghi2012design}) at lower pre-load, which could be a fruitful research direction.

To conclude, additive manufacturing of very soft TPEs is challenging, but the prospect of adding soft silicone rubber-like materials in the filament/pellet-based MEX process makes it worthwhile. Furthermore, the printed membranes are simple yet demonstrate the feasibility of using soft TPEs for soft fluidic actuators and/or sensors. Moreover, the ballooning of TPEs can be utilized by soft roboticists to achieve significant expansion in a soft package. Thus, 3D-printed soft TPEs seem a promising class of materials for soft robot applications.

%%%%%%%%%%%%%%%%%%%%%%%%%%%%%%%%%%%%%%%%%%%%%%%%%%%%%%%%%%%%%%%%%%%%%%%%%%%%%%%%%%%%%%%%%%%%%%%%%%
\section{Materials and Methods}

\subsection{3D Printing of Thermoplastic Elastomers}
We used Styrene-Ethylene-Butadiene-Styrene (SEBS) G1657 (Kraton Corporation, USA) and Supersoft TF3ZGO-LCNT TPE (Kraiburg TPE, Germany) pellets with a Shore Hardness of 47A and 00-30, respectively. We printed both pellets with a 0.4 mm nozzle with the SEBS at 230 $^{\circ}$C and the Supersoft at 195 $^{\circ}$C. Other printing parameters are discussed in the Supplementary Information.

\subsection{Uniaxial Testing}
To characterize the tensile behavior, we used an Instron 3343 tensile tester (Instron, USA) with a speed of 80 mm/min. In this tensile test, we used dogbones with a geometry based on the ISO37 standard type 2 \cite{brown2018physical}. We printed the TPE dogbones (SEBS and Supersoft) with the lines infill (Figure \ref{fig:fig1}(f)). In addition, we used a 3D-printed mold to cast three soft silicone rubbers: Dragon Skin 30, Eco-Flex 00-30, and Eco-Flex 00-10 (Smooth-on, Inc., USA). Lastly, we printed Ninjaflex (Ninjatek, USA) dogbones as an example of a flexible filament. All dogbones were fabricated in triplicate. After fabrication, their cross-section was measured to scale the force to an engineering stress. 

\subsection{Inflation Experiments}
To characterize the inflation behavior, we printed two samples of each membrane. Subsequently, we placed the membranes in our inflation setup (Figure \ref{fig:Fig2}(a)) for characterization. In this setup, we tracked the inflation of the SEBS membranes through MATLAB (The MathWorks, USA), while also recording the pressure and curvature using an Arduino Uno (Arduino AG, Italy) with an MXP5001DP pressure sensor (NXP Semiconductors, The Netherlands) and a camera, respectively. A manually triggered solenoid valve connected to a regulated pressure source turned the pressure input on or off. The experimental protocol consisted of three main steps: (I) set the pressure using the pressure regulator, (II) start the measurements and open the valve, and (III) close the valve after 30 seconds. We repeated the protocol from (I) three times before increasing the pressure. After reaching the ballooning pressure, we restarted from the lowest pressure. Afterwards, we cropped the inflated membrane at the maximum expansion to evaluate the stretch by fitting an ellipse and used its arc length to compute the stretch. The characterization of the Supersoft membranes was slightly different due to the lower pressures (below 10 kPa). In this experiment, we manually increased the pressure while recording the pressure and a video. Subsequently, we reconstructed the Supersoft membrane expansion over time using the algorithm described in the Supplementary Information.

\subsection{Cyclic Testing of Membranes}
The Supersoft and SEBS membranes (three and six layers) were printed and placed in the inflation setup. These membranes were exposed to up to 1,000 inflation/deflation cycles. Each cycle lasted eight seconds with 1.5 seconds of inflation and 6.5 seconds of deflation. The pressure magnitudes were set to (three-layer/six-layer) 5.7/4.75 and (average) 44/80 kPa for the Supersoft and SEBS membranes, respectively. An Arduino Uno was used to control the inflation/deflation cycle using a three-way valve, while the pressure was measured using an MPX5100DP (NXP Semiconductors, The Netherlands). 

\subsection{Bending Actuator}
We reused the pressure control of the inflation experiments to actuate the bending actuator. To connect the pressure source to the actuator, a silicone tube and ferrule were added. The ferrule was inserted into the printed part while heated using a soldering iron to ensure a secure fit. We added stripes to the bending actuator to track the curvature using a camera, which we used to construct a circle in MATLAB to estimate the curvature. In addition, force experiments were performed using the Instron tensile tester in “Hold mode” over a 25-30-second span, which was repeated three times at the same pressure. Before every force experiment, we tared the load cell.

\subsection{Membraned Sucker and Membrane-Based Gripper Characterization}
Both membrane-based actuators used a setup built around an Arduino Rev4 Minima (Arduino AG, Italy) to benefit from its higher analog-to-digital resolution, which was connected to an MPXH6400AC6T1 pressure sensor (NXP Semiconductors, The Netherlands). We tested the adhesion force using the Instron tensile tester at 100 mm/min with and without muscle gel (Spirosan, The Netherlands) applied to the soft foot as a mucus with a pre-load of 1 N. In addition, we experimented with pre-loads ranging from 1 to 25 N without the gel. In both force experiments, the measurement was repeated three times. Furthermore, we characterized the tactile fluidic sensing using the Instron tensile tester with probes with a 3, 6, and 12 mm diameter, which moved down for 5 mm at 50 mm/min for ten cycles.

\bibliographystyle{ieeetr}
\bibliography{references}

\end{document}